\documentclass{article}

\usepackage[preprint,nonatbib]{neurips_2020}
\usepackage[utf8]{inputenc} 
\usepackage[T1]{fontenc}    
\usepackage{hyperref}       
\usepackage{url}            
\usepackage{booktabs}       
\usepackage{amsfonts}       
\usepackage{nicefrac}       
\usepackage{microtype}      
\usepackage{amsmath,amssymb} 
\usepackage{subfigure}
\usepackage{graphicx}
\usepackage{authblk}
\DeclareMathOperator*{\argmin}{\arg\min}

\title{Dropout Inference with Non-Uniform Weight Scaling}

\author[1]{%
  \textbf{Zhaoyuan Yang}}

\author[1]{%
  \textbf{Arpit Jain}}

\affil[1]{GE Research,  Niskayuna \& San Ramon, US}
\affil[]{ \{zhaoyuan.yang, arpit.jain\}@ge.com}

\begin{document}

\maketitle
\begin{abstract}
Dropout as regularization has been used extensively to prevent overfitting for training neural networks. During training, units and their connections are randomly dropped, which could be considered as sampling many different submodels from the original model. At test time, weight scaling and Monte Carlo approximation are two widely applied approaches to approximate the outputs. Both approaches work well practically when all submodels are low-bias complex learners. However, in this work, we demonstrate scenarios where some submodels behave closer to high-bias models and a non-uniform weight scaling is a better approximation for inference.

\end{abstract}

\section{Introduction}
\label{sec:intro}
Neural networks are becoming increasing bigger, deeper, and more complex. Large number of parameters make deep learning models prone to overfitting especially with limited numbers of data. Dropout \cite{hinton2012improving,JMLR:v15:srivastava14a} as regularization has been used extensively to prevent overfitting for training neural networks with limited data. During training, units and their connections are randomly dropped, which could be considered as sampling many different submodels from the original model. At test time, weight scaling and Monte Carlo approximation are two widely used approaches to approximate the outputs \cite{JMLR:v15:srivastava14a,gal2016dropout}. Both approaches work well for many problems since dropout is similar to bagging \cite{breiman1996bagging} which combines exponential number of complex learners \cite{warde2013empirical}. 

In this work, we start with reviewing related work and background of dropout, then we demonstrate scenarios where some submodels, trained with dropout, could be high-bias models, and a non-uniform weight scaling, similar to boosting \cite{schapire1990strength}, is a better approximation of outputs. 


\section{Related Works}

Dropout was first proposed by Hinton, et. al. \cite{hinton2012improving} to prevent overfitting for training neural networks. There has been substantial empirical and theoretical work in understanding mechanism of dropout\cite{labach2019survey}. This work focuses on standard dropout at inference time; thus, in the subsequent sections, the discussion is concentrated on dropout inference. 

Weight scaling and Monte Carlo approximation \cite{JMLR:v15:srivastava14a,gal2016dropout} are two widely applied approaches to approximate the dropout outputs. During test time, weight scaling uses a scaled-down version of the trained weights while Monte Carlo approximation samples different submodels and performs model averaging. Optimal choices of inference approximation are problem dependent \cite{goodfellow2016deep}. Compared to weight scaling, Monte Carlo approximation is more expensive for evaluation, but Monte Carlo approximation could provide an estimation for uncertainty \cite{gal2016dropout}. To reduce computation for Monte Carlo approximation, dropout distillation \cite{bulo2016dropout} was proposed to provide an approximation for Monte Carlo approximation.  

Both weight scaling and Monte Carlo approximation weigh each sub-network equally, which are similar to the bagging method \cite{breiman1996bagging}. Warde, et. al. \cite{warde2013empirical} performed an empirical analysis of dropout bagging versus dropout boosting through bagging and boosting large number of low-bias complex models, they discover that dropout boosting performs no better than dropout bagging. The intuition is that dropout is an ensemble of complex learners; thus, bagging is better than boosting when all individual base models are complex learners. 

In this work, we demonstrate scenarios which are different than the experiments shown in \cite{warde2013empirical}. Instead of ensembling individual networks, we obtain a single network with dropout using training and validation data. At test time, instead of uniformly weighing each submodels (weight scaling and Monte Carlo approximation), we use a non-uniform weight scaling vector determined by constrained optimization. This is more closed to boosting than bagging since each submodels are weighted non-uniformly. We show that, in some situations, non-uniform weight scaling is a better approximation for inference than uniform weight scaling and Monte Carlo approximation due to biases in some submodels.

\section{Background}
This section briefly describes the standard dropout models \cite{JMLR:v15:srivastava14a,gal2016dropout} and some dropout inference methods which are related to our work. 

\subsection{Dropout for Training}
We use bold upper-case letters to represent matrices, bold lower-case to represent vectors and $\odot$ to represent Hadamard (element-wise) product. Consider a neural network with $L$ hidden layers, and let $i$ index the hidden layer of the network. Let $\mathbf{z}^{i}$ denote the input vector and $\mathbf{z}^{i+1}$ denote the output vector for layer $i$. For layer $i$, a weight matrix is represented as $\mathbf{W}^i$, a bias vector is represented as $\mathbf{b}^i$ and an activation function is represented as $f_i$.  The standard neural network output for layer $i$ can be described as 
\begin{align*}
	\mathbf{z}^{i+1} = f_i(\mathbf{W}^i \mathbf{z}^{i} + \mathbf{b}^i)
\end{align*}
\label{eq:1}
With the dropout operation, the network, during training time, can be described as
\begin{align*}
	\mathbf{z}^{i+1} = f_i(\mathbf{W}^i (\mathbf{d}^i\odot \mathbf{z}^{i}) + \mathbf{b}^i)
\end{align*}
\label{eq:1}
where $\mathbf{d}^i = [d_1^i, d_2^i, ..., d_n^i]^\top$ is a binary random vector such that $d_x^i \sim \text{Bernoulli}(p)$ with probability $p$ being 1, and $n$ being dimension of the input vector $\mathbf{z}^{i}$. 

With a slight abuse of notation, let $\boldsymbol{\theta}$ denote the parameters of the neural network, $\mathbf{d}$ denote the concatenate binary random vector from each layer with dropout, $\mathcal{L}$ denote object function for training, $\mathbf{X}^{(train)}$ and $\mathbf{y}^{(train)}$ represent training data and their corresponding labels. The objective of training can be described as

\begin{align*}
	\boldsymbol{\theta}^* = \argmin_{\boldsymbol{\theta}}\mathbb{E}_{\mathbf{d} \sim \text{Bernoulli}(p)} \mathcal{L}(\boldsymbol{\theta}, \mathbf{d}; \mathbf{X}^{(train)}, \mathbf{y}^{(train)})
\end{align*}

Let $F$ denote the function parametrized by learnt parameters $\boldsymbol{\theta}$ which takes an input vector $\mathbf{x}$, a binary vector $\mathbf{d}$ and generate a prediction $\hat{y}$. At inference time, the outputs can be described as

\begin{align*}
	\hat{y} =\mathbb{E}_{\mathbf{d} \sim \text{Bern}(p)} F(\mathbf{x}, \mathbf{d}; \boldsymbol{\theta}) \tag{1}
\end{align*}
\label{eq:2}
However, there are $2^n$ combinations for binary random vector $\mathbf{d}$ which makes evaluation computationally expensive. Weight scaling and Monte Carlo approximation are two popular approaches to approximate the dropout inference output.
\subsection{Weight Scaling}
Instead of evaluating $\mathbb{E}(f(x))$, weight scaling \cite{JMLR:v15:srivastava14a} approximate the output by $f(\mathbb{E}(x))$. In most of cases, $f(\mathbb{E}(x)) \neq \mathbb{E}(f(x))$, but it works well practically. Weight scaling for layer $i$ can be described as
\begin{align*}
	\mathbf{z}^{i+1} = f(\mathbb{E}_{\mathbf{d}\sim \text{Bern}(p)}(\mathbf{W}^i (\mathbf{d}^i \odot \mathbf{z}^i) + \mathbf{b}^i))
	   = f(\mathbf{W}^i (\mathbb{E}_{\mathbf{d} \sim \text{Bern}(p)}(\mathbf{d}^i) \odot \mathbf{z}^i) + \mathbf{b}^i)
\end{align*}
since $\mathbb{E}_{d \sim \text{Bern}(p)}(d) = p$, the expression can be simplifid as  
\begin{align*}
	\mathbf{z}^{i+1} = f(p\mathbf{W}^i \mathbf{z}^i + \mathbf{b}^i)
\end{align*}
where $p\mathbf{W}^i$ can be considered as scaled-down version of trained weights. If activation function $f$ is linear, weight scaling outputs will be the same as (1). For a single layer network with dropout and softmax activation, weight scaling is equivalent as taking the geometric mean of all submodels \cite{goodfellow2016deep}.

\subsection{Monte Carlo Approximation}
Instead of scaling down the trained weight, Monte Carlo approximation \cite{JMLR:v15:srivastava14a,gal2016dropout} samples large number of binary random vectors $\mathbf{d}_k$, and approximates the output through arithmetic mean or geometric mean. In certain problems, Monte Carlo approximation, given enough samples, could provide better inference outputs. Let $N$ denote number of sampling iterations, a Monte Carlo approximation output can be described as
\begin{align*}
	\hat{y} = \frac{1}{N} \sum_{k=1}^{N} F( \mathbf{x}, \mathbf{d}_k; \boldsymbol{\theta}) \tag{arithmetic mean}
\end{align*}

\begin{align*}
	\hat{y} =\left(\prod _{k=1}^{N}F( \mathbf{x}, \mathbf{d}_k; \boldsymbol{\theta})\right)^{\frac {1}{N}} \tag{geometric mean}
\end{align*}
 
Both arithmetic mean and geometric mean are widely applied approaches to obtain approximation outputs. Choices are problem dependent. 

\subsection{Bagging and Boosting}

Bagging (bootstrap aggregation) \cite{breiman1996bagging} and boosting \cite{schapire1990strength} are two popular ensemble methods. They are trained with different objectives. Bagging works well with complex learners (low-bias) while boosting works well with weak learners (high bias, slightly better than random). Let $M$ denote number of models in the ensemble, where each base model is represented by $h_m(\mathbf{x})$ and $F_{ensemble}(\mathbf{x})$ be the function of the ensemble, an example of bagging methods can be obtained by averaging over all the base models uniformly
\begin{align*}
	F_{ensemble}(\mathbf{x}) = \frac{1}{M} \sum_{m=1}^{M} h_m(\mathbf{x}) \tag{bagging}
\end{align*}
while each model in boosting methods usually associates with a unique weight value. Let $\alpha_m$ represent the weight value of model $h_m(\mathbf{x})$, an example of boosting methods is given by
\begin{align*}
	F_{ensemble}(\mathbf{x}) = \frac{1}{M} \sum_{m=1}^{M} \alpha_m h_m(\mathbf{x}) \tag{boosting}
\end{align*}

Even though dropout training and inference are similar to bagging, they are not exactly the same. For example, each base model in bagging is trained independently while all the submodels trained with dropout share parameters \cite{warde2013empirical}.

\section{Non-uniform Weight Scaling for Dropout Inference}
Both weight scaling and Monte Carlo approximation are similar to the bagging method since they weight each submodels equally. This approximation works well when all submodels are low-bias strong learners \cite{warde2013empirical}. 

However, for some problems, we conjecture that the low-bias assumption in all submodels may not hold due to limited data, fast convergence, limited sampling submodels, early stopping, etc. For these problems, some submodels may overfit to the data while others are still under-fit to the data, and weight each submodel uniformly could be suboptimal. We demonstrate such phenonmenon through a non-uniform weight scaling formulation described in the following section.
\subsection{Non-uniform Weight Scaling} 
For the layer $i$ during inference time, weight scaling scales down weights uniformly by probability $p$. We propose a different formulation where weights are scaled non-uniformly during inference time. We introduce a scale vector $\mathbf{s} = [s_1, s_2, ...., s_n]^\top$ which has the same dimension as the input vector $\mathbf{z}^i$, and the inference output can be described as
\begin{align*}
	\mathbf{z}^{i+1} = f(\mathbf{W}^i (\mathbf{s}\odot \mathbf{z}^i) + \mathbf{b}^i)
\end{align*}

There are some additional constraints for the scale vector $\mathbf{s}$ which can be described as
\begin{align*}
	\frac{1}{n} \sum_{k=1}^{n} s_k = p \text{ and } 0 \leq s_k \leq 1
\end{align*}

Weight scaling is a special case of our scale vector where all scale values are uniform. 
Consider a network (parametrized by $\boldsymbol{\theta}$) with only single dropout operation, finding the optimal scale vector $\mathbf{s}$ can be formulated as a constrained optimization problem

\begin{align*}
\min_{\mathbf{s}} \: & \mathcal{L}(\mathbf{s}; \mathbf{X}^{(train)}, \mathbf{y}^{(train)}, \boldsymbol{\theta}) \\
s.t.  \: & \frac{1}{n} \sum_{k=1}^{n} s_k = p \text{ and } 0 \leq s_k \leq 1
\end{align*}
\label{eq:1}

In implementation, we use reparametrization, Lagrangian and KKT conditions to determine the optimal scale vector $\mathbf{s}^*$. We introduce a new vector $\mathbf{e} = [e_1, e_2, ..., e_n]^\top$ and $\mathbf{s}$ can be reparametrization by $\mathbf{e}$ as
\begin{align*}
\mathbf{s} = \mathbf{e} - \frac{1}{n} \sum_{k=1}^{n} e_k + p \tag{2}
\end{align*}
then we always have $\frac{1}{n} \sum_{k=1}^{n} s_k = p$, and formulation of the optimization becomes
\begin{align*}
\mathbf{s}^* = \argmin_{\mathbf{s}} \: & \mathcal{L}(\mathbf{s};\mathbf{X}^{(train)}, \mathbf{Y}^{(train)}, \boldsymbol{\theta}) + \sum_{k=1}^{n} \lambda_k \left(\max(0, s_k - 1) - \min(0, s_k)\right) \tag{3}\\
\end{align*}
where $\mathbf{s}$ is reparametrization by $\mathbf{e}$ as shown in (2) and $\lambda_k$ is a large positive number penalize infeasible values. Compare with (uniform) weight scaling, non-uniform weight scaling is more similar to boosting than bagging. Similar to weight scaling, non-uniform weight scaling can be merged into weight matrix without adding any new parameters.

\section{Experiments and Implementation}

In this section, we show experiment results and implementation details. Our focus is on demonstrating a phenomenon, but not on pushing the state-of-the-art results, thus, we intentionally use simple setups. 

\subsection{Training Base Models}

We use CIFAR-10 \cite{krizhevsky2009learning} and Fashion-MNIST \cite{xiao2017fashion} to demonstrate our observation. For both datasets, we split training data into 80\% training and 20\% validation. We do not have any data augmentation for Fashion-MNIST. For CIFAR-10, we use random flip and random translation (3 pixels). We build small networks with residual connection \cite{he2016deep} across convolutional layers (figure 4). Dropout is placed between the GAP layer and output layer. Dropout rate is set to 0.5 in our experiments. We use SGD with learning rate 0.01 and momentum 0.9 for optimization. For CIFAR-10, we train the model with batch size of 32 for 72 epoch. For Fashion-MNIST, we train the model with batch size of 32 for 64 epoch. These numbers are determined based on performance on the validation data (validation accuracy does not improve significantly with large number of epoch) and early stopping. Models are saved based on performance on the validation data (using weight scaling) during end of each epoch. Once training finished, we freeze all parameters before determining the scale vector for weight scaling.

\subsection{Non-uniform Weight Scaling for Combining Submodels}
Experiments are performed in Tensorflow \cite{tensorflow2015-whitepaper}. Instead of scaling the outputs after dropout at inference time, Tensorflow scales the outputs after dropout during training time. Thus, for a dropout rate of 0.5, constraints for the scale vector $\mathbf{s}$ implemented by Tensorflow should be
\begin{align*}
	\frac{1}{n} \sum_{k=1}^{n} s_k = 1 \text{ and } 0 \leq s_k \leq 2
\end{align*}

We optimize the scale vector $\mathbf{s}$ based on the training data using Adam optimizer (learning rate is 0.001, $\beta_1$ is 0.9 and $\beta_2$ is 0.999). We observe there is no significant difference between Adam (learning rate 0.001) and SGD (learning rate 0.01) for optimizing the scale vector in terms of performance. We set $\lambda$ in (3) to 10000 and optimal scale vectors are determined using validation data.   

We repeat experiments 8 times for each dataset with different training-validation split and record the performances of weight scaling, Monte Carlo approximation and non-uniform weight scaling. In our experiments, weight scaling can be approximated as geometric mean of averaging all the submodels since we use softmax as output activation and dropout is placed before the output layer (similar as EfficientNet \cite{tan2019efficientnet}, but we do not have dropout between convolutional layers). 

We summarize the validation error and testing error for CIFAR-10 in Table 1 and figure 1. The results for Fashion-MNIST are summarized in Table 2 and figure 2. We discover a consistent improvement after non-uniform weight scaling compare with uniform weight scaling (geometric mean) and Monte Carlo approximation (arithmetic mean) which is different with results shown in previous work \cite{warde2013empirical}. Figure 3 shows an example distribution of scale vectors, we observe that most of scale vectors are not centralized around 1, thus, we conjecture that, due to early stopping and limited data, some submodels are over-fit to the training data while other submodels are under-fit. Consequently, weight each submodel uniformly is suboptimal. 

\begin{table}[!h]
\begin{center}
\begin{tabular}{llll}
\toprule
{} &  Uniform Scaling & Monte Carlo &  Non-uniform Scaling \\
\midrule
val error (\%)  &    9.66 $\pm$ 0.18 &  9.69 $\pm$ 0.19 &  \textbf{9.41} $\pm$ 0.21 \\
test error (\%) &    9.98 $\pm$ 0.20 &  9.98 $\pm$ 0.22 &  \textbf{9.83} $\pm$ 0.25 \\
\bottomrule
\end{tabular}
\caption{Validation error (\%) and test error (\%) on CIFAR-10 dataset. Mean and standard deviation are calculated based on 8 different experiments.}
\end{center}
\end{table}

\begin{table}[!h]
\begin{center}
\begin{tabular}{llll}
\toprule
{} & Uniform Scaling & Monte Carlo  &   Non-uniform Scaling \\
\midrule
val error (\%)  &    7.72 $\pm$ 0.15 &  7.73 $\pm$ 0.16 &  \textbf{7.53} $\pm$ 0.16 \\
test error (\%)  &    8.37 $\pm$ 0.25 &  8.37 $\pm$ 0.25 &  \textbf{8.18} $\pm$ 0.23 \\
\bottomrule
\end{tabular}
\caption{Validation error (\%) and test error (\%) on Fashion-MNIST dataset. Mean and standard deviation are calculated based on 8 different experiments.}
\end{center}
\end{table}

\begin{figure}[!h]
	\centering
	\includegraphics[scale=0.35]{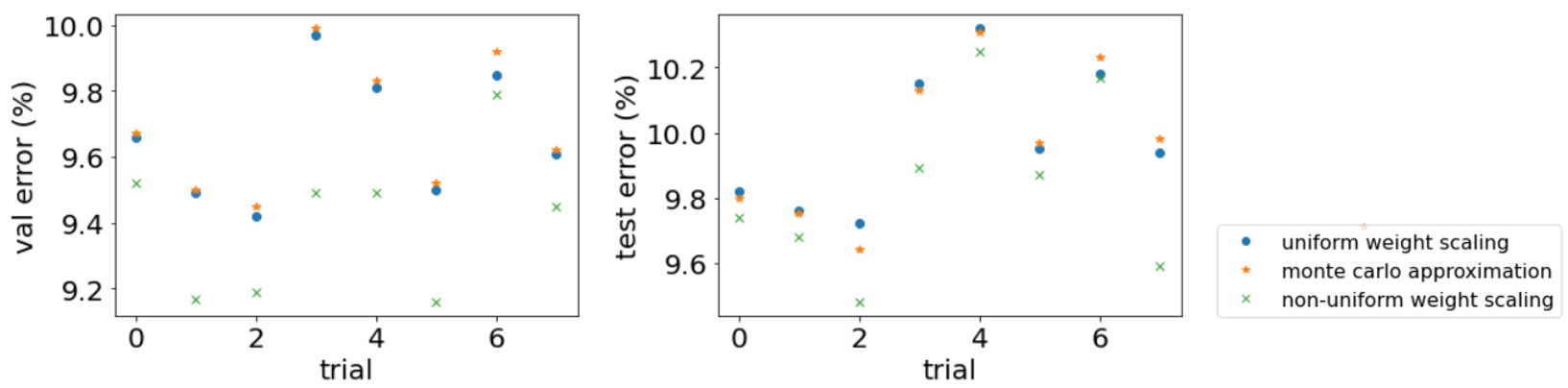}
	\caption{Validation error (left) and test error (right) on CIFAR-10 dataset for 8 different experiments.}
	\label{fig:cifar10}
\end{figure}

\begin{figure}[!h]
	\centering
	\includegraphics[scale=0.35]{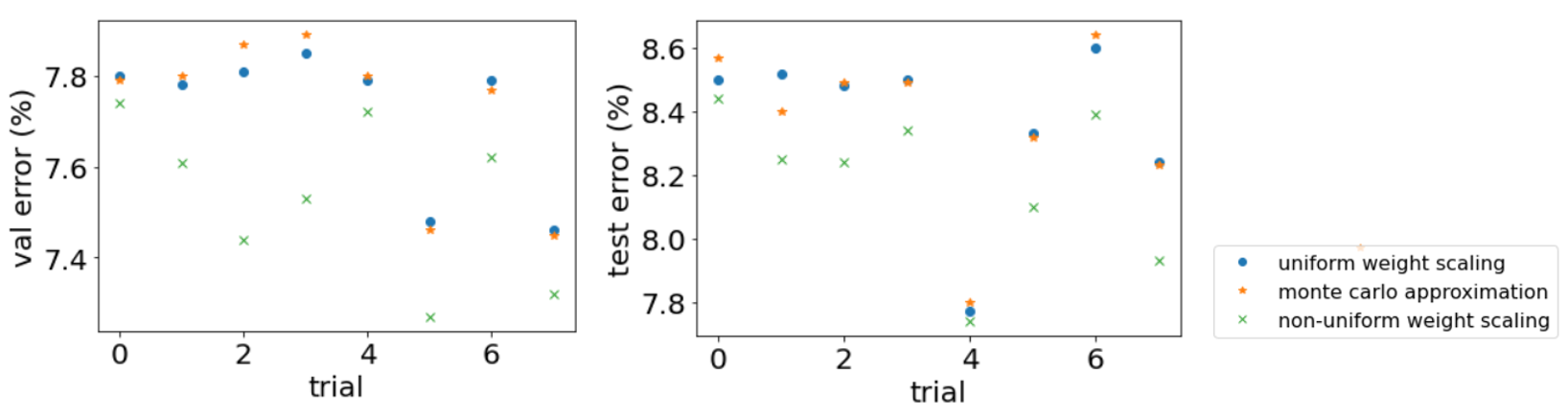}
	\caption{Validation error (left) and test error (right) on Fashion-MNIST dataset for 8 different experiments.}
	\label{fig:fashion_mnist}
\end{figure}

\begin{figure}[!h]
	\centering
	\includegraphics[scale=0.5]{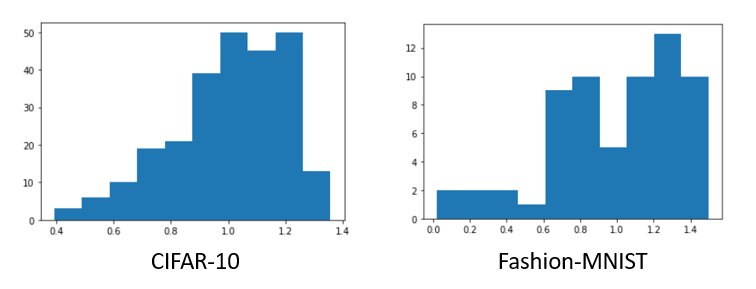}
	\caption{An example distribution of scale vector from two experiments.}
	\label{fig:scaling_vector}
\end{figure}


\begin{figure}
    \centering
    \subfigure[][]{
	\label{CIFAR-10-architecture}
    \includegraphics[width=4.3cm]{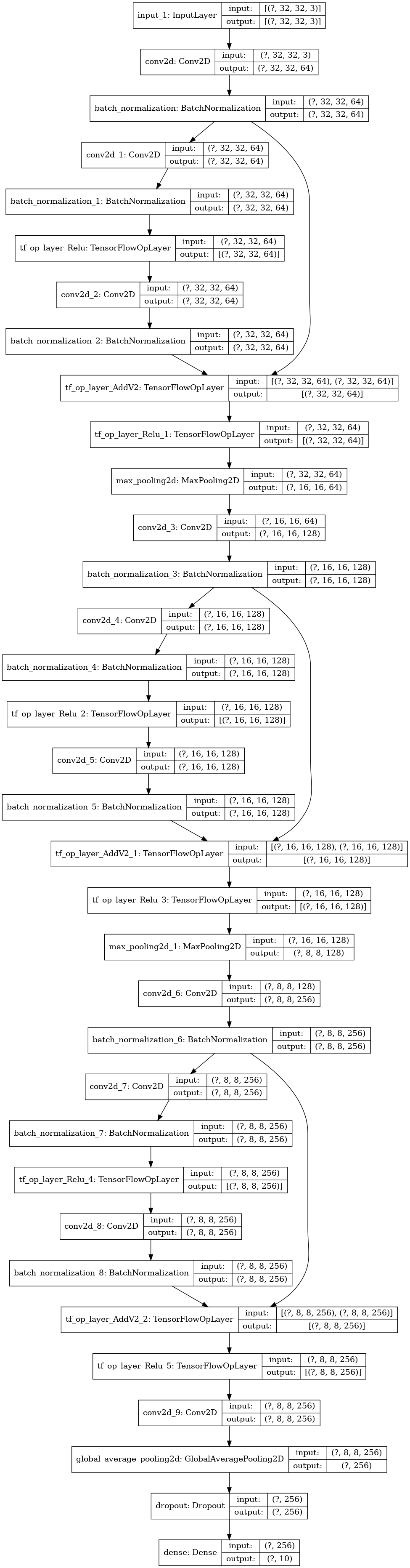}}
    \hspace{8pt}
	\subfigure[][]{
    \label{Fashion-MNIST-architecture}
    \includegraphics[width=6cm]{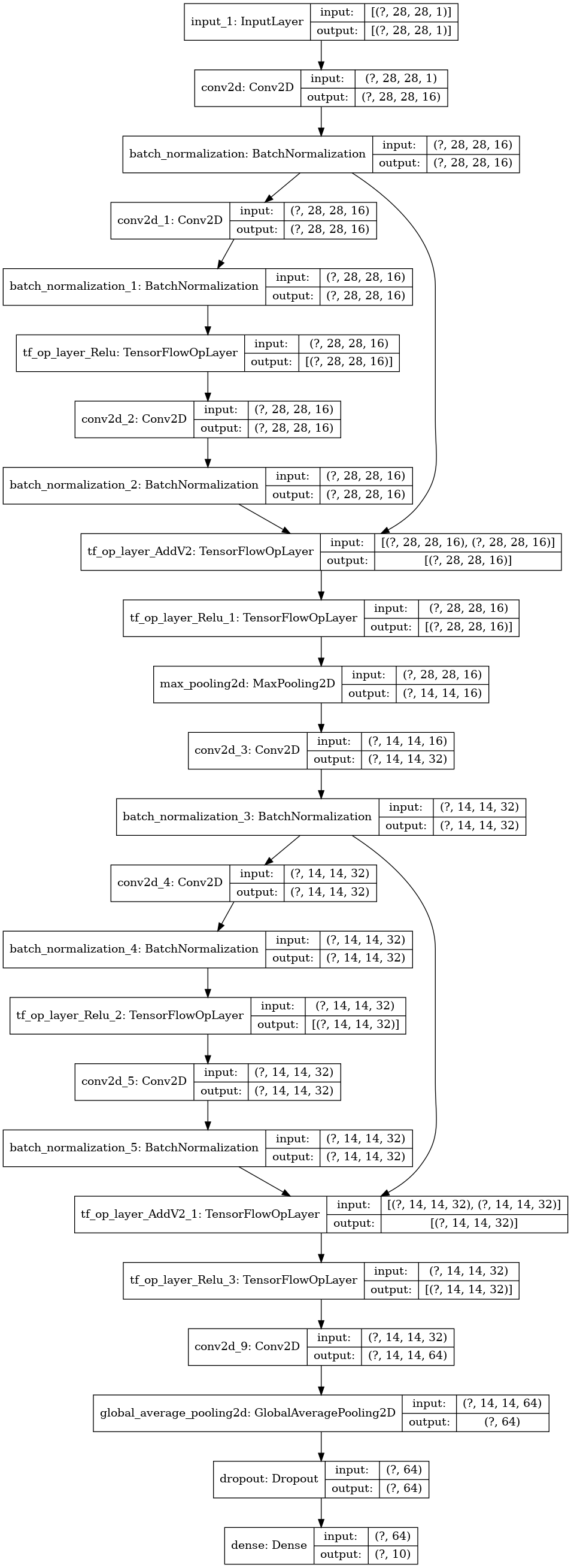}}
    \caption{Architecture of models used in the experiment section. (a) Architecture for CIFAR-10. (b) Architecture for Fashion-MNIST. Kernel sizes for both models are 3x3.}
    \label{fig:mode_arch}
\end{figure}

\section{Insights and Future Work}

In this section, we describe the insights we learnt from experiments and potential future work. During experiments with different hyper-parameters and networks, we discover that if the optimal performance on validation data occurs at an early epoch number, the selected model is likely consisted of both low-bias submodels as well as high-bias submodels. In this situation, non-uniform weight scaling may be better than (uniform) weight scaling and Monte Carlo approximation, which is contradicted to the observation in previous work \cite{warde2013empirical}. In some early stopping scenarios, non-uniform weight scaling may also bring some additional improvements. If a model is selected when its training performance is nearly perfect (e.g. optimal validation performance occurs when training performance is nearly perfect), we discover that non-uniform weight scaling does not provide additional improvement (due to overfit issues). We also discover that when a model takes long epoch number to reach the optimal performance on validation data, it is difficult to tell whether uniform weight scaling or non-uniform weight scaling could provide a better performance. Optimal choices seem to depend on the problems and selection criteria of the optimal model. For future work, we plan to investigate impacts of non-uniform weight scaling with various hyperparameters such as different network architectures, learning rate schedulers, batch sizes, etc. We know that non-uniform weight scaling scales each submodels differently which is similar to boosting, but we don't know whether it assigns more weights on low-bias models or it uses sequence of high-bias models to reduce the overall bias. We plan to investigate this as well in our future work.

\section{Conclusion}
We empirically discover that, for some scenarios, models trained with dropout behave more similar to boosting than bagging which is different than results shown in previous work \cite{warde2013empirical}. We conjecture that this phenonmenon is caused by over-fitting of some submodels on limited data while other submodels are still under-fitting. In this situation, the original model is consisted of both low-bias submodels and high-bias submodels, we propose a non-uniform weight scaling and observe that a non-uniform weight scaling could provide a better result for such situation. 
\bibliographystyle{unsrt}
\bibliography{egbib}

\begin{thebibliography}{10}

\bibitem{hinton2012improving}
Geoffrey~E Hinton, Nitish Srivastava, Alex Krizhevsky, Ilya Sutskever, and
  Ruslan~R Salakhutdinov.
\newblock Improving neural networks by preventing co-adaptation of feature
  detectors.
\newblock {\em arXiv preprint arXiv:1207.0580}, 2012.

\bibitem{JMLR:v15:srivastava14a}
Nitish Srivastava, Geoffrey Hinton, Alex Krizhevsky, Ilya Sutskever, and Ruslan
  Salakhutdinov.
\newblock Dropout: A simple way to prevent neural networks from overfitting.
\newblock {\em Journal of Machine Learning Research}, 15(56):1929--1958, 2014.

\bibitem{gal2016dropout}
Yarin Gal and Zoubin Ghahramani.
\newblock Dropout as a bayesian approximation: Representing model uncertainty
  in deep learning.
\newblock In {\em international conference on machine learning}, pages
  1050--1059. PMLR, 2016.

\bibitem{breiman1996bagging}
Leo Breiman.
\newblock Bagging predictors.
\newblock {\em Machine learning}, 24(2):123--140, 1996.

\bibitem{warde2013empirical}
David Warde-Farley, Ian~J Goodfellow, Aaron Courville, and Yoshua Bengio.
\newblock An empirical analysis of dropout in piecewise linear networks.
\newblock {\em arXiv preprint arXiv:1312.6197}, 2013.

\bibitem{schapire1990strength}
Robert~E Schapire.
\newblock The strength of weak learnability.
\newblock {\em Machine learning}, 5(2):197--227, 1990.

\bibitem{labach2019survey}
Alex Labach, Hojjat Salehinejad, and Shahrokh Valaee.
\newblock Survey of dropout methods for deep neural networks.
\newblock {\em arXiv preprint arXiv:1904.13310}, 2019.

\bibitem{goodfellow2016deep}
Ian Goodfellow, Yoshua Bengio, and Aaron Courville.
\newblock {\em Deep learning}.
\newblock MIT press, 2016.

\bibitem{bulo2016dropout}
Samuel~Rota Bul{\`o}, Lorenzo Porzi, and Peter Kontschieder.
\newblock Dropout distillation.
\newblock In {\em International Conference on Machine Learning}, pages 99--107.
  PMLR, 2016.

\bibitem{krizhevsky2009learning}
Alex Krizhevsky, Geoffrey Hinton, et~al.
\newblock Learning multiple layers of features from tiny images.
\newblock 2009.

\bibitem{xiao2017fashion}
Han Xiao, Kashif Rasul, and Roland Vollgraf.
\newblock Fashion-mnist: a novel image dataset for benchmarking machine
  learning algorithms.
\newblock {\em arXiv preprint arXiv:1708.07747}, 2017.

\bibitem{he2016deep}
Kaiming He, Xiangyu Zhang, Shaoqing Ren, and Jian Sun.
\newblock Deep residual learning for image recognition.
\newblock In {\em Proceedings of the IEEE conference on computer vision and
  pattern recognition}, pages 770--778, 2016.

\bibitem{tensorflow2015-whitepaper}
Mart\'{i}n Abadi, Ashish Agarwal, Paul Barham, Eugene Brevdo, Zhifeng Chen,
  Craig Citro, Greg~S. Corrado, Andy Davis, Jeffrey Dean, Matthieu Devin,
  Sanjay Ghemawat, Ian Goodfellow, Andrew Harp, Geoffrey Irving, Michael Isard,
  Yangqing Jia, Rafal Jozefowicz, Lukasz Kaiser, Manjunath Kudlur, Josh
  Levenberg, Dandelion Man\'{e}, Rajat Monga, Sherry Moore, Derek Murray, Chris
  Olah, Mike Schuster, Jonathon Shlens, Benoit Steiner, Ilya Sutskever, Kunal
  Talwar, Paul Tucker, Vincent Vanhoucke, Vijay Vasudevan, Fernanda Vi\'{e}gas,
  Oriol Vinyals, Pete Warden, Martin Wattenberg, Martin Wicke, Yuan Yu, and
  Xiaoqiang Zheng.
\newblock {TensorFlow}: Large-scale machine learning on heterogeneous systems,
  2015.
\newblock Software available from tensorflow.org.

\bibitem{tan2019efficientnet}
Mingxing Tan and Quoc Le.
\newblock Efficientnet: Rethinking model scaling for convolutional neural
  networks.
\newblock In {\em International conference on machine learning}, pages
  6105--6114. PMLR, 2019.

\end{thebibliography}
\end{document}